\documentclass[conference]{IEEEtran}
\IEEEoverridecommandlockouts
\usepackage{cite}
\usepackage{amsmath,amssymb,amsfonts}
\usepackage{algorithmic}
\usepackage{graphicx}
\usepackage{textcomp}
\usepackage{xcolor}
\usepackage{booktabs}
\usepackage{enumerate}
\usepackage{multirow}
\def\BibTeX{{\rm B\kern-.05em{\sc i\kern-.025em b}\kern-.08em
    T\kern-.1667em\lower.7ex\hbox{E}\kern-.125emX}}

\begin{document}

\title{Rapid Detection of Aircrafts in Satellite Imagery based on Deep Neural Networks\\
{\footnotesize \textsuperscript{}}
\thanks{\** Research Center For Modeling and Simulation (RCMS), NUST
\newline \vspace{5mm}
}
}

\author{\IEEEauthorblockN{\textsuperscript{} Arsalan Tahir}
\IEEEauthorblockA{\text{\** RCMS, NUST} \\
\text{Islamabad, Pakistan} \\
atahir.mscse17@rcms.nust.edu.pk}
\and

\and
\IEEEauthorblockN{\textsuperscript{} Muhammad Adil}
\IEEEauthorblockA{\text{RCMS, NUST} \\
\text{Islamabad, Pakistan}\\
madil.mscse17@rcms.nust.edu.pk}
\and

\IEEEauthorblockN{\textsuperscript{} Arslan Ali}
\IEEEauthorblockA{\text{RCMS, NUST} \\
\text{Islamabad, Pakistan}\\
aali.msbi17@rcms.nust.edu.pk}

}

\maketitle

\begin{abstract}

Object detection is one of the fundamental objectives in Applied Computer Vision. In some of the applications, object detection becomes very challenging such as in the case of satellite image processing. Satellite image processing has remained the focus of researchers in domains of Precision Agriculture, Climate Change, Disaster Management, etc. Therefore, object detection in satellite imagery is one of the most researched problems in this domain. This paper focuses on aircraft detection. in satellite imagery using deep learning techniques. In this paper, we used YOLO deep learning framework for aircraft detection. This method uses satellite images collected by different sources as learning for the model to perform detection. Object detection in satellite images is mostly complex because objects have many variations, types, poses, sizes, complex and dense background. YOLO has some limitations for small size objects (less than$\sim$32 pixels per object), therefore we upsample the prediction grid to reduce the coarseness of the model and to accurately detect the densely clustered objects. The improved model shows good accuracy and performance on different unknown images having small, rotating, and dense objects to meet the requirements in
real-time.
\end{abstract}


\section{Introduction}
Object detection is a fundamental challenge in computer vision and also a key part of active research. The purpose of object detection is to find an instance in a specific location by drawing a bounding box \cite{liu2018deep}. Many high-level vision tasks are also solvable using object detection like segmentation, activity recognition and event capturing. Traditional machine learning techniques are not suitable to perform in realtime for object detection. With invent of deep learning, the computer is capable to understand the visual imagery like a human. In satellite imagery, object detection is a very complicated task due to low-resolution pixel and have densely clustered objects. In deep learning three frameworks which give solution for object detection are Faster RCNN \cite{ren2015faster}, YOLO \cite{redmon2016you} and SSD \cite{liu2015ssd}. YOLO has the greatest inference speed and score on the Pascal VOC dataset \cite{everingham2010pascal}. The main problem is to locate aircraft from the large searching area of the image in an efficient manner.

\begin{figure}[htbp]

\centerline{\includegraphics[width=3.3in, height=2.2in]{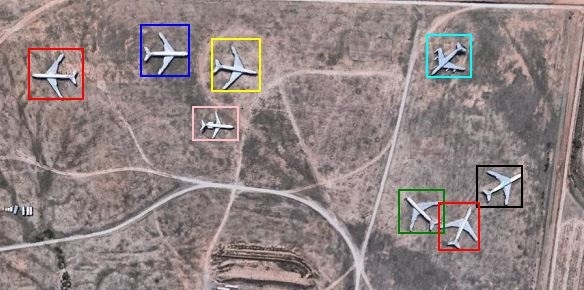}}
\caption{Bounding boxes in above image show the detection of aircrafts. }
\label{fig1}
\end{figure}
For this purpose, a complete autonomous UAVs application are required, which used classification and localization in real-time. Authors used YOLO based deep neural network to solve this problem. This paper presents object detection of an aircraft for satellite images based on YOLO real-time detector.

This paper is divided into six sections. First section covers the introductory part and the second section covers related work after the introduction. After the second section network architecture for object detection is discussed. The methodology and results are covered in the fourth and fifth section and conclusion are highlighted in the last section.

\section{Related Work}
The frequently used techniques of object detection suggested by national and international academics are split primarily into three categorizations: based on motion data, based on the extraction of features, and based on the matching of templates. Qinhan et al. \cite{luo2016airplane} used multiple windows that have the highest chances of objects and then apply SVM and HOG techniques for the generation of proposals, but the disadvantage of it used fixed-size windows. Cheng et al.  \cite{cheng2009detecting} used subtraction and registration techniques for the identification of objects in satellite images. Lee et al.  \cite{lee2017real} applied RCNN for the detection of objects in images taken by UAVs. Azevedo et al.  \cite{azevedo2014automatic} used median background techniques for the identification of objects in aerial imagery. J. Khan et al. \cite{khan2017automatic} proposed automated target detection for satellite images using edge boxes algorithm. Junyan Lu et al. \cite{lu2018vehicle} propose a method for the detection of vehicles using YOLO deep learning framework. Douillard \cite{site2} proposed a deep learning method for detection of an object in satellite images. They used RetinaNet architecture on COCW dataset based on Faster RCNN. Lu Zhang et al. \cite{zhang2015hierarchical} present a hierarchical oil tank detector with deep surrounding features for high-resolution optical satellite imagery. The proposed method is divided into three modules named candidate selection, feature extraction and classification. Marcum et al. \cite{marcum2017rapid} propose a method to localize the surface to air-missile (SAM) sites using sliding window approach for satellite images. If the object is hundreds of meters then this approach performs better results rather than small objects which is computationally expensive. For small objects, millions of sliding window cutouts generate over 10-meter area in Digital Globe image.
With the arrival of deep learning and GPU technology, the development becomes fast and efficient in the field of computer vision especially when we are solving problems of pattern recognition and image processing and these are more robust rather than traditional techniques. Deep learning techniques plays a very important part in the field of object detection because it can extract features from an image automatically. Deep learning provides excellent accuracy in the field of object detection. If we want to detect category (a horse, a cat, a dog) then first collect the large dataset and start training on this dataset. After training when we give the image for prediction and then output is produced in the form of vector scores for each category. We define an objective function that calculates the error between desired output and vector output score. For this computation machine sets internal parameters (weights or real number) to minimize the error. In deep learning, millions of weights are used for training. So the gradient vector is used for each weight vector and tells what amount of error is decreases and increases. The gradient vector is then adjusted in the opposite direction of the weighted vector. So many practitioners use stochastic gradient descent to minimizing the loss function with randomly \cite{sherrah2016fully} \cite{qu2017vehicle} \cite{he2015spatial}. Therefore this paper uses YOLO deep learning method to obtain real-time detection and performance in satellite images.

\section{Network Architecture}
Redmon et.al proposed a method YOLO \cite{redmon2016you}, is a real-time object detector based on a convolutional neural network. After some time Joseph Redmon and Ali Farhadi released a new version YOLOv2, which has good performance and speed \cite{redmon2017yolo9000}. Now the latest version is YOLOv3 proposed by Joseph Redmon and Ali Farhadi with the increment of layers in architecture to improve speed and accuracy \cite{redmon2018yolov3}. YOLO has many advantages rather than traditional algorithms because of its architecture. The traditional method used region proposal networks for the generation of proposals and then implement the CNN on these proposals for feature extraction. These methods are slow and not real-time due to their two-stage detection architecture for satellite imagery. YOLO takes an image with resolution 416 × 416 × 3 and divides the input image into S × S grid. If the center of the object falls into the grid then that grid is responsible for predicting the object. Each cell in grid predicts the bounding boxes (B) and confidence scores related to those boxes. YOLO v1 has a large positioning mistake and low recall rate compared to the region-based proposal technique such as Fast R-CNN. The primary improvements of YOLOv2 are therefore to improve the rate of recall, batch normalization, anchor boxes, and multiscale training. Batch normalization is a popular method to normalize the data at the time of training and also used to increase the speed with mean 0 and variance 1, which can prevent the gradient descent for vanishing. Batch normalization also helps to make network convergence faster. Faster RCNN used to add fully connected layers to predict bounding boxes directly after the convolutional layers but YOLO uses anchor boxes, which improves the speed and recall rate.

\begin{figure*}[htbp]

\centerline{\includegraphics[width=7in]{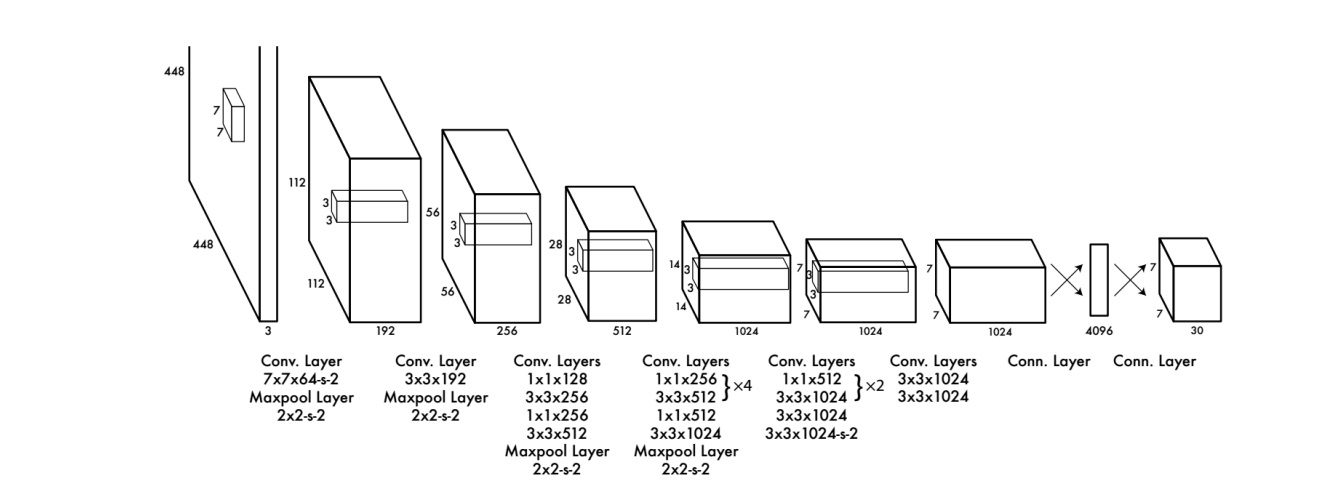}}
\caption{YOLO CNN Architecture}
\label{fig2}
\end{figure*}

During training, YOLO adjusts the input after every 10 epoch to make the model performed well at the test time on the multiscale images. The CNN architecture using in this paper has 24 convolutional layers followed by two fully connected layers. To reduce the coarseness model uses prediction grid is 26 $\times 26 $ and downsample the factor by 16. This architecture shows the highest speed and accuracy. Figure 2 displays the whole architecture of CNN and also preferred due to computational speed and accuracy. The final layer is used for the classification of objects with probability in between 0 and 1.
\section{Dataset}
First, we collect datasets from DigitalGlobe satellite and apply some preprocessing and
data augmentation techniques. In preprocessing we converted all image in 550 × 350
resolution to reduce the training time. After that we used labeling tools for tagging of
images. After tagging we converted the data into standard architecture using python
language.Datasets have played very important role in the area of object detection. It is most
important factor for measuring and analysis of performance of different algorithms and
also pushing this field towards challenging and complex problems. The internet makes
it possible to capture diversity and richness of objects in large images with large number
of categories. The increase in large scale datasets with millions of images has played
important role and opened unprecedented performance in object detection.

The classifiers show poor results on satellite images due to the effect of different conditions.
\begin{itemize}
\item Spatial Resolution
  \begin{itemize}
   \item Objects are very small and densely clustered in satellite images rather than the prominent and large object and for small objects like cars, the object is only \textasciitilde15 pixels in high-resolution images.
   \end{itemize}

\item Rotation Invariance
   \begin{itemize}
   \item Objects in satellite imagery have many orientations (for example ships have any orientation ranging from 0 to 360 degree).
   \end{itemize}

\item Training example frequency
   \begin{itemize}
   \item There is relative dearth of data in satellite imagery and objects are not clearly visible in shape.
   \end{itemize}

\item	Ultra-high resolution
   \begin{itemize}
   \item Images are of very high resolution (hundreds of megapixels) but most algorithms take input images with few hundreds of pixels.  Upsampling the image means object of interest becomes large, dispersed and not feasible for standard architecture and downsampling the image can change the object shape.
   \end{itemize}

\item  Temporal (time of day/season/year)

   \begin{itemize}
   \item Seasonal differences and time of day also effect on satellite images.
   \end{itemize}

\end{itemize}

Therefore it is difficult for the classifier to detect objects from conventional datasets due to mentioned reasons on satellite images. For this, we need a specialized kind of data for satellite images for the processing which is computationally less expensive and time-efficient. Here the some datasets, which are using for Aerial images. VEDAI Dataset Razakarivony et.al \cite{razakarivony2016vehicle} made a dataset VEDAI (Vehicle Detection in Aerial Images) collected from public Utah ARGC database. The images have three RGB channel and one infrared channel. The authors split the images into 1024 × 1024 RGB channel and perform downsampling to convert the images into 512 × 512 pixels and ignore the infrared channel. The authors used just RGB channels and also set the GSD (Ground Sample Distance) is 12.5 cm. This dataset consist of nine vehicle classes and total images are 1250 (“plane”, “boat”,
“camping car”, “tractor”, “van”, “pick- up” and “other”). The annotation of images has five parts: Object class and four coordinates of objects. Mundhenk et al. \cite{mundhenk2016large} made a dataset COWC(Cars Overhead with Context) and collected from six different locations. The image size of images is 2000 × 2000 pixels and the total number of images is 53 with the format of TIFF. They covered areas of six locations namely Columbus, Utah (United States), Selwyn (New Zealand), Postdam (Germany), Tornoto (Canada) and Vaihingen. The images of Columbus and Vaihingen are in grayscale while remaining are in the RGB channel. The object size in the image is 24 pixel with GSD of 15 cm per pixel. They annotate the 32,716 images with car object and annotation includes object class and four coordinates of objects. DOTA Dataset Guisong et al.  \cite{xia2018dota} made a dataset DOTA (Dataset for object detection in Aerial images) of aerial images and collected from different sources like google earth an airplane or ship. We used Bbox Labeling tool for tagging of aircraft. and sensors. The GSD of images is diversified and characterized by multiresolution and multi-sensor. DOTA images are 4000 ×4000 pixels and classes are 15 with annotations namely (“plane”, “storage tank”, “swimming pool”, “ship”, “harbor”, “bridge”, “helicopter”, and “other”). The annotation of images has five parts: Object class and four coordinates of objects. All above mentioned datasets belong to aerial imagery.
\newline
There are many reasons behind the creation of dataset. First the dataset of satellite imagery are not commonly available. Second two or three datasets are available and those datasets
have less number of objects. For this purpose, we collected images from DigitalGlobe and convert according for standard architectures by applying some preprocessing techniques. Data annotation is process of labelling the data of specific instance like human, car etc. which is understandable for machines. Data annotation is performed manually by human using the annotation tool and stored large amount of data for machine learning.
The area of objects is cropped through bounding boxes and coordinates of objects are stored in file for learning of machines. We used open-source Bounding Box Label tool for ground truth boxes of aircrafts in the dataset \cite{site}.

\section{Methodology}
We performed two steps in methodology in which, first we make a dataset for standard architecture and second configure the parameters for training to obtain results. We collect a dataset of satellite images from different sources and manually annotate the images and draw anchor boxes on desired objects. There are two parts in the dataset: images, which are in JPEG format and labels, which are in text format. Evert text file is saved according to images, which contain the annotation of objects and the format of the annotation is:

\begin{equation}
\begin{aligned}
  <object-class> <x,y,w,h>
\end{aligned}
\end{equation}

Where x and y are the center points of object and w and h are the width and height of object correspondence to the image and name of object class. The input dimension of YOLO is 416 × 416 ×3 for training but you should care about the image size should not large may lose the useful information. The basic information of publically available datasets of aerial imagery is described in Table II.

We process our dataset and convert in the form of standard architecture using:
\newline
\begin{itemize}

\item  Center points

\begin{equation}
\begin{aligned}
  x=(x_{max}+x_{min})/2\
\end{aligned}
\end{equation}

\begin{equation}
\begin{aligned}
  y=(y_{max}+y_{min})/2\
\end{aligned}
\end{equation}

\item  Width and Height

\begin{equation}
\begin{aligned}
  w=(x_{max}+x_{min})\
\end{aligned}
\end{equation}

\begin{equation}
\begin{aligned}
  h=(y_{max}+y_{min})\
\end{aligned}
\end{equation}
\end{itemize}

The batch training passes the dataset through learning algorithms and save the weights. Batch size represents the training examples in one forward pass. The learning rate is used for optimization and minimizing the loss function of neural network and loss function maps values of variables onto real number and also show the associated cost with values. During training of neural network it is common to use decay because after each update weights are multiplied by value less than 1 and also prevents the weights from growing too large. Momentum is used to improve both training speed and accuracy. Our network consist of 26 × 26 grid and was tested on one object class and returned 26× 26 × 11 tensor. We used batch\_size =64 and filter =30 for training.
\begin{figure}[htbp]

\centerline{\includegraphics[width=3.5in,height=2.4in]{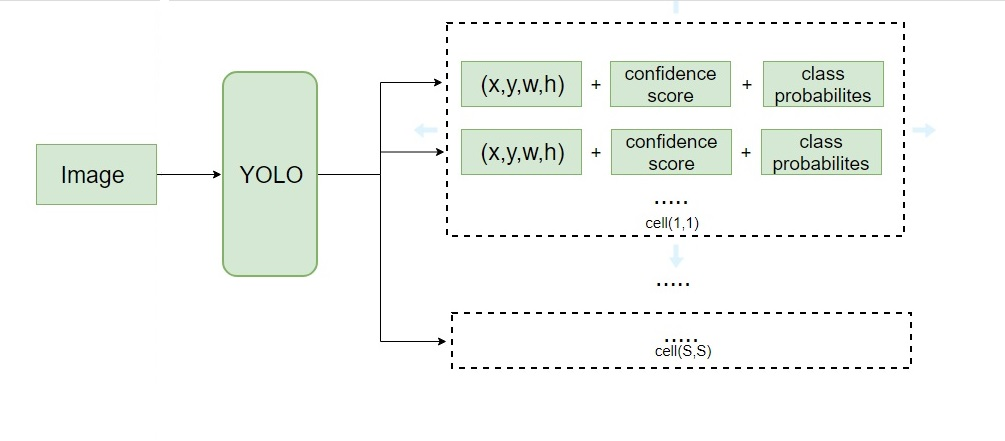}}
\caption{Testing output of YOLO  }
\label{fig3}
\end{figure}

 Figure \ref{fig3} shows YOLO generates number of bounding boxes when the input images is given and it use non max suppression technique to find the correct bounding boxes around the object with maximum intersection over Union.

\section{Results}
\begin{figure}[htbp]

\centerline{\includegraphics[width=3.4in,height=2.2in]{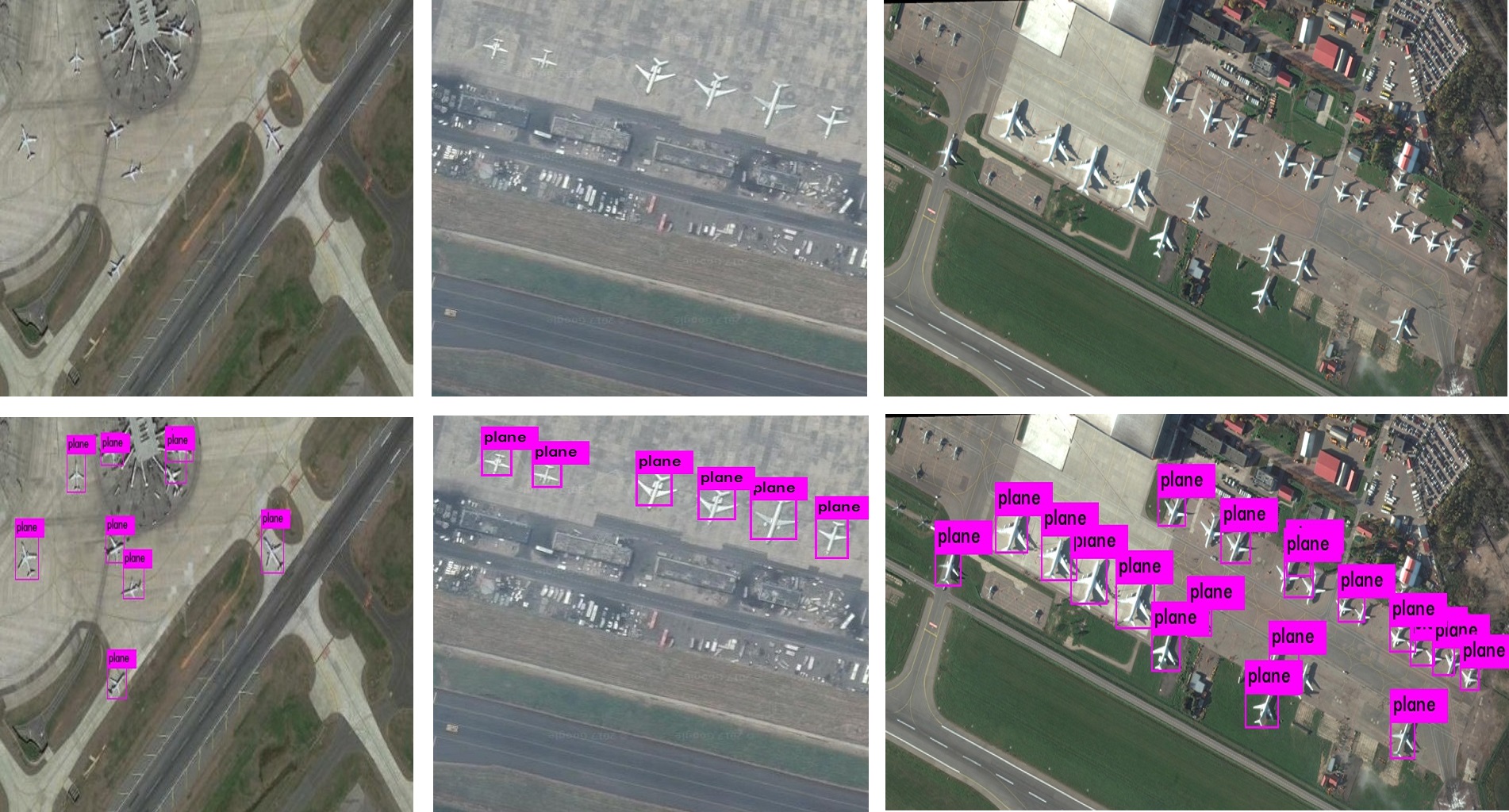}}
\caption{Test Results on Unseen Images}
\label{fig4}
\end{figure}

In this paper we used NVIDIA GPU Geforce GTX 1060 for training. Authors changed the architecture of the model according to object size ($\sim$10 pixels per object) and performed training on the custom dataset. The detection results of unknown images are shown in figure \ref{fig4}. Figure \ref{fig4} (left) shows that our model has good results and perform well on small objects. In middle figure also shows good results but in Figure \ref{fig4} (right) there is one object, which is not detected. But the overall model gives good results and also detects objects within milliseconds.  Our model detects more than 96 \% of aircrafts. Results showed in table \ref{tab:addscore} that YOLO was able to identify “aircraft” objects in the dataset with 90.20\% accuracy.
\begin{table}[htbp]
  \centering
  \caption{Test results on unseen testing images}
  \scalebox{1.2}{
    \begin{tabular}{|c|c|c|c|c|}
    \toprule
    Indicator & accuracy   & Precision & F1-score   & fps \\
    \midrule
    value & 94.20\%  & 99\%  &96 \%  & 55 \\
    \bottomrule
    \end{tabular}%

  \label{tab:addscore}%
  }
\end{table}%

\begin{table*}[htbp]
  \centering
  \caption{Comparison with other Satellite Imagery (Aircraft) datasets}
  \scalebox{1.6}{
    \begin{tabular}{|c|c|c|c|c|c|}
    \toprule
    \textbf{Sr.} & \textbf{Name} & \textbf{Number Of Objects} & \textbf{Type} & \textbf{Description} \\
    \midrule
    1 & NWPU-RESISC45 Dataset    & 700     & Aircraft & (class,0,1) \\
    \midrule
    2  & NWPU VHR-10 Dataset     & 800     & Aircraft & (class,0,1) \\
    \midrule
    3  & Custom Dataset    & 2213   & Aircraft & (class,x,y,w,h) \\
    \bottomrule
    \end{tabular}%
  \label{tab:adddbinfo}%
  }
\end{table*}%

\section{Conclusion}
In this paper rapid aircraft detection based on YOLO deep learning is presented. We used 2200 objects for training with tuning the parameters to calculate the anchors for a good intersection over the union. The improved model has good results on unknown images of small and densely clustered objects and also meet the real-time requirements. Results show this approach is fast and robust for aircraft detection in dense airports. Next, we will increase the number of objects and classes to achieve good performance and accuracy.

\section*{Conflicts of Interests}
The author of this paper shows no conflicts of interest.

\bibliographystyle{IEEEtran}
\bibliography{bib}

\end{document}